\documentclass[sigconf,nonacm]{acmart}
\usepackage{booktabs}  % 负责 \toprule, \midrule, \bottomrule (三线表)
\usepackage{multirow}  % 负责 \multirow (跨行合并)
\usepackage{graphicx}  % 负责 \resizebox 和 \scalebox (表格缩放)
\usepackage{stfloats}
\AtBeginDocument{%
  }

\renewcommand\footnotetextcopyrightpermission[1]{}
\acmConference[ ]{ }{ }{ }
\acmBooktitle{ }
\acmPrice{ }
\acmISBN{ }

\begin{document}
\sloppy
\title[HST-HGN: Spatio-Temporal Hypergraph for Fatigue Detection]{HST-HGN: Heterogeneous Spatial-Temporal Hypergraph Networks with Bidirectional State Space Models for Global Fatigue Assessment}

\author{Changdao Chen}
\email{ccd2236114971@stu.xjtu.edu.cn}
\orcid{1234-5678-9012}
\affiliation{%
  \institution{School of Computer Science and Technology, Xi’an
Jiaotong University}
  \city{Xi'an}
  \state{Shaanxi}
  \country{China}
}

\author{Qinqiuhong Ye}
\email{uzumymw@stu.xjtu.edu.cn}
\affiliation{%
  \institution{School Of Mechanical Engineering, Xi'an Jiaotong University}
  \city{Xi'an}
  \state{Shaanxi}
  \country{China}
}

\author{Hao Chen}
\email{chen1580129807@stu.xjtu.edu.cn}
\affiliation{%
  \institution{Xi'an Jiaotong University Health Science Center}
  \city{Xi'an}
  \state{Shaanxi}
  \country{China}
}

\author{Jinyu Wang}
\email{jinyu.wang@xjtu.edu.cn}
\affiliation{%
  \institution{School of Computer Science and Technology, Xi'an Jiaotong University}
  \city{Xi'an}
  \state{Shaanxi}
  \country{China}
}

% \affiliation{%

% \affiliation{%

% \affiliation{%

% \affiliation{%

% \affiliation{%

\renewcommand{\shortauthors}{}

\begin{abstract}
It remains challenging to assess driver fatigue from untrimmed videos under constrained computational budgets, due to the difficulty of modeling long-range temporal dependencies in subtle facial expressions. Some existing approaches rely on computationally heavy architectures, whereas others employ traditional lightweight pairwise graph networks, despite their limited capacity to model high-order synergies and global temporal context. Therefore, we propose HST-HGN, a novel Heterogeneous Spatial-Temporal Hypergraph Network driven by Bidirectional State Space Models. Spatially, we introduce a hierarchical hypergraph network to fuse pose-disentangled geometric topologies with multi-modal texture patches dynamically. This formulation encapsulates high-order synergistic facial deformations, effectively overcoming the limitations of conventional methods. In temporal terms, a Bi-Mamba module with linear complexity is applied to perform bidirectional sequence modeling. This explicit temporal-evolution filtering enables the network to distinguish highly ambiguous transient actions, such as yawning versus speaking, while encompassing their complete physiological lifecycles. Extensive evaluations across diverse fatigue benchmarks demonstrate that HST-HGN achieves state-of-the-art performance. In particular, our method strikes a balance between discriminative power and computational efficiency, making it well-suited for real-time in-cabin edge deployment.
\end{abstract}


%    <concept>
%        <concept_id>10010147.10010178.10010224.10010225.10010228</concept_id>
%        <concept_desc>Computing methodologies~Activity recognition and understanding</concept_desc>
%        <concept_significance>500</concept_significance>
%        </concept>
%    <concept>
%        <concept_id>10010147.10010257.10010293.10010294</concept_id>
%        <concept_desc>Computing methodologies~Neural networks</concept_desc>
%        <concept_significance>300</concept_significance>
%        </concept>
%    <concept>
%        <concept_id>10010405.10010481.10010485</concept_id>
%        <concept_desc>Applied computing~Transportation</concept_desc>
%        <concept_significance>100</concept_significance>
%        </concept>
%  </ccs2012>

% \ccsdesc[500]{Computing methodologies~Activity recognition and understanding}
% \ccsdesc[300]{Computing methodologies~Neural networks}
% \ccsdesc[100]{Applied computing~Transportation}
\keywords{Driver Fatigue Detection, Hypergraph Neural Networks, State Space Models, Multimodal Fusion, Spatial-Temporal Modeling}

\maketitle
\begin{figure}[t] 
  \centering
  \includegraphics[width=\linewidth]{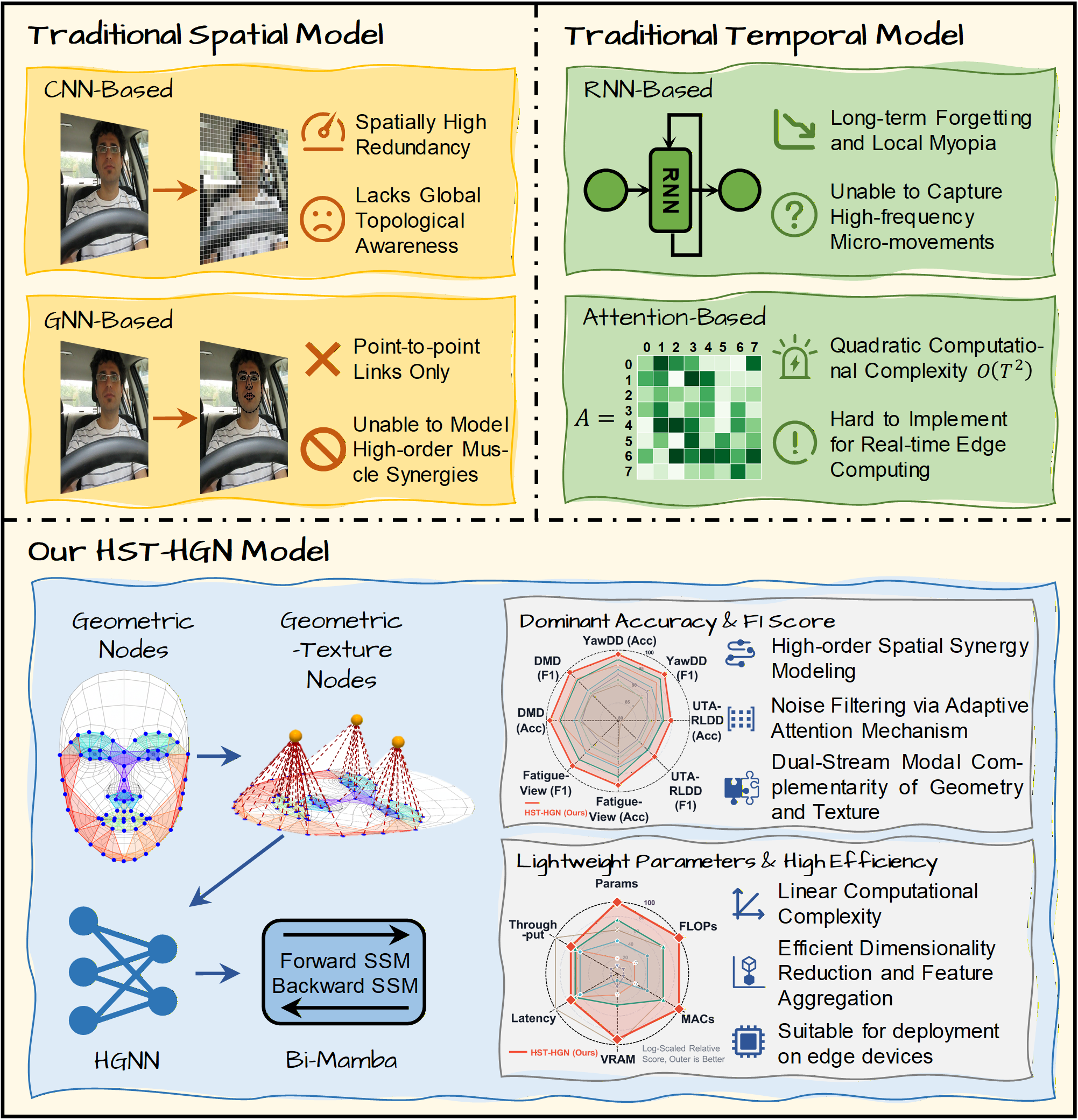} 
  \caption{Qualitative comparison between traditional spatial/temporal methods and our HST-HGN.}
  \label{fig:teaser}
\end{figure}
\section{Introduction}

Fine-grained facial dynamic analysis is a cornerstone of human-computer interaction (HCI) . With its assistance, information from raw videos containing faces can be effectively extracted, thereby providing new solutions to challenging problems like driver fatigue assessment. Fatigued driving poses a significant threat to the safety of both drivers and pedestrians.\cite{10517310}. However, distinct from explicit action recognition, driver fatigue—characterized by behaviors like recurrent yawning and microsleep—is difficult to detect and typically deepen gradually over extended periods.\cite{elnabi2024machine,8482470}. It takes a novel paradigm, which is capable of capturing subtle local facial deformations and encoding long-range temporal dependencies to accurately estimating global fatigue states and pinpointing critical temporal segments from untrimmed video sequences.

To this end, extensive work has explored diverse spatial-temporal modeling strategies. Spatially, appearance-based CNNs remain the dominant paradigm for extracting discriminative texture cues such as eye closure and mouth opening from RGB frames\cite{SUN2023106981, zhao2024review}. Geometry-driven GCNs extend this line by representing facial landmarks as graph nodes linked via anatomical or kinematic priors, improving robustness to pose variation and occlusion while reducing input redundancy\cite{FA2023103826, WU2025111106}. Temporally, sliding-window RNN/LSTM pipelines are widely adopted to encode the evolution of blinking and yawning patterns over short segments\cite{10018105, ZHANG2024124124}. More recently, studies have shifted toward Transformer to directly encode feature or explicit fatigue parameter sequences, exploiting global self-attention to uncover long-range dependencies across extended temporal horizons\cite{hassan2025realtime}.

Nevertheless, these established paradigms encounter inherent bottlenecks in global fatigue assessment. Spatially, CNN-based backbones are computationally heavy and remain sensitive to large head poses and motion blur\cite{zhao2024review}. While GCN-driven models alleviate efficiency concerns, they often struggle to perceive critical texture semantics—such as eye-closure status—leading to sub-optimal multimodal fusion\cite{elnabi2024machine, WU2025111106}. Temporally, sliding-window-based RNNs/LSTMs suffer from local myopia and long-term forgetting, failing to capture the global context of extended sequences\cite{10159554,li2024videomambastatespacemodel,8953343}. Furthermore, although Transformers excel in global modeling, their quadratic computational complexity, $O(T^2)$, impedes efficient deployment on resource-constrained edge devices for long video processing\cite{SAHA2025130417,pmlr-v139-bertasius21a}. %Consequently, existing frameworks frequently fail to differentiate highly ambiguous facial behaviors, such as yawning and speaking, which exhibit similar spatial structural perturbations but possess markedly distinct temporal frequencies.

To overcome these bottlenecks, we propose HST-HGN, a novel heterogeneous spatial-temporal hypergraph architecture driven by bidirectional State Space Models for robust and efficient global fatigue assessment. Rather than dense local windows, a global sparse sampling strategy is adopted to extract heterogeneous representations from extended sequences. Spatially, we introduce a 3D canonical alignment mechanism integrated with a heterogeneous super-node topology. Through star-topology hyperedges, our model broadcast local texture semantics into the global geometric framework without causing dimensionality explosion. Temporally, Bi-Mamba layers are applied instead of self-attention mechanism. Leveraging selective state spaces, HST-HGN achieves global temporal context modeling with linear complexity, effectively capturing the frequency differences between yawning and speaking. % effectively capturing both the low-frequency evolution of yawning and the high-frequency oscillations of speaking. 
%Finally, a global temporal max-pooling mechanism is employed to implicitly localize salient fatigue segments, preserving critical high-frequency transients.

The primary contributions of this work are summarized as follows:
\begin{itemize}
\item We propose HST-HGN, an efficient global fatigue assessment framework that pioneers the integration of heterogeneous hypergraphs with Bi-SSMs, enabling robust long-range video modeling with linear complexity.
\item We design a novel hierarchical super-node topology featuring 3D canonical alignment, which effectively decouples rigid pose variations and facilitates the high-order fusion of local texture semantics and global geometric structures via hyperedges.
\item We demonstrate the inherent interpretability of our framework, enabling weakly supervised temporal localization of fatigue events and significantly mitigating action ambiguity between behaviors like speaking and yawning.
\item Extensive evaluations on benchmark datasets validate that HST-HGN demonstrates state-of-the-art (SOTA) performance in both accuracy and computational efficiency, highlighting its substantial potential for real-world edge deployment.
\end{itemize}
%%%% 时态用现在时
%%%% 已改{begin}
\section{Related Work}
\subsection{Spatial Representation for Facial Behavior}
Robust spatial representation is the foundation for assessing driver fatigue. Conventional deep learning paradigms primarily leverage 2D convolutional neural networks, such as ResNet and VGG, to extract dense texture features from facial regions of interest (ROIs) \cite{SUN2023106981,xxxxx,minhas2022smart,han2025dense}. To capture short-term motion, architectures such as C3D and I3D extended these convolutions into the temporal domain\cite{tran2015learning,carreira2017quo}. Nevertheless, such grid-based models impose a prohibitive computational burden for continuous edge deployment and exhibit great sensitivity to unconstrained environments, suffering from severe feature degradation under head-pose variations and illumination shifts\cite{9185000,wijnands2020realtime}.

To overcome the computational bottlenecks of dense pixel processing, Graph Convolutional Networks (GCNs) emerged to model the non-Euclidean topologies of structural joints. Building upon foundational skeleton-based architectures like ST-GCN \cite{9557757} and 2s-AGCN, dynamic GCNs have been further adapted to model facial topologies over structural landmarks \cite{FA2023103826} and relational patches \cite{JIANG2023119640}. However, standard GCNs remain restricted to pairwise edges, making it hard to encapsulate high-order synergistic deformations of facial muscle groups during physiological expressions such as yawning \cite{FA2023103826}. To break these pairwise restrictions, spatio-temporal hypergraphs have recently emerged to capture high-order structural correlations in fine-grained physiological tasks, such as motor symptom assessment \cite{An2025ASH} and spasms detection \cite{Wang2026MSTHGCNAM}. While recent advances further explore multi-modal hypergraph fusion \cite{e26030239} and Transformer-integrated temporal reasoning \cite{Wang2024HyperSTTNSG}, combining high-order hypergraphs with self-attention inevitably incurs quadratic computational overhead. This prohibits their deployment on edge devices for long-video processing, thereby motivating our exploration of a more efficient temporal engine.

\subsection{Global Temporal Modeling \& SSMs}
%In untrimmed videos, effectively modeling long-range temporal dependencies is essential for distinguishing high-frequency actions, such as speaking, from low-frequency physiological states like prolonged yawning. 
To distinguish the frequency difference between talking and yawning from untrimmed videos, it is essential to model long-range temporal dependencies. Historically, Recurrent Neural Networks—particularly LSTMs—have been extensively employed to aggregate sequential frame features through gating mechanisms\cite{9185000,al2023lstm,cao2025optimized}. However, when processing long sequences, LSTMs suffer from gradient vanishing during Backpropagation Through Time (BPTT), causing severe long-term forgetting. As a remedial paradigm, Vision Transformers (ViTs) and their video-centric variants, such as TimeSformer, leverage global self-attention to establish unfettered long-range dependencies\cite{Wang2023Selective,li2024videomambastatespacemodel,10.1007/978-3-031-19833-5_6}. 
Regrettably, the computational complexity of the attention matrix grows quadratically with sequence length $T$, which is unacceptable on edge devices. More recently, State Space Models (SSMs) have surfaced as a transformative alternative for sequence modeling. The S4 layer parameterizes continuous-time dynamics using HiPPO projections, enabling efficient linear-time handling of long sequences and serving as a backbone in long-form video models such as ViS4mer and S5-based selective token frameworks\cite{10.1007/978-3-031-19833-5_6,Wang2023Selective,somvanshi2025s4mambacomprehensivesurvey}. Based on S4, Mamba introduces a hardware-aware selective scan algorithm, allowing the model to dynamically filter noise while retaining critical contextual memory, thereby achieving a global receptive field with linear complexity, $O(T)$\cite{2025Rethinking,Tang2024VMRNN,somvanshi2025s4mambacomprehensivesurvey}.  VideoMamba further adapts this architecture to video by scanning spatial‑temporal tokens with forward–backward selective SSMs\cite{li2024videomambastatespacemodel}.
%%% 已改{end}

%%% 已改{begin}
\section{Methodology}
\begin{figure*}[t] 
  \centering
  \includegraphics[width=\textwidth]{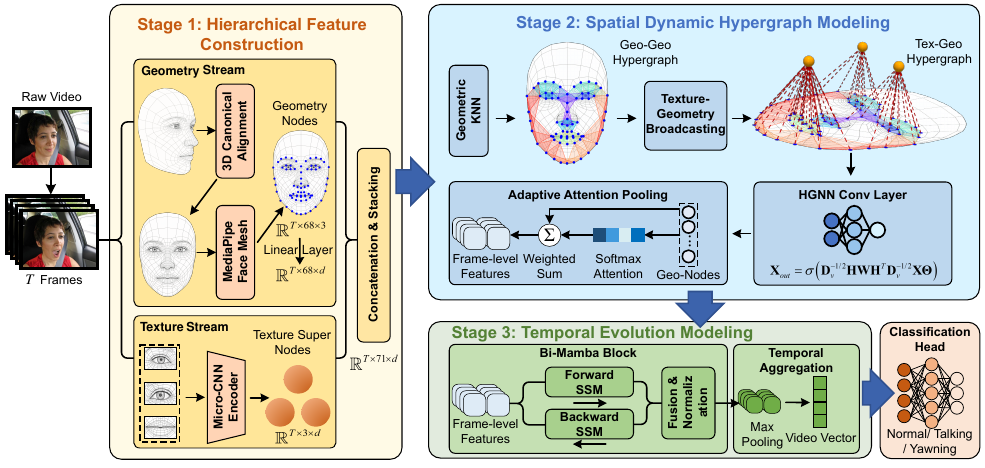} 
  \caption{An overview of HST-HGN framework. The architecture consists of three core stages. Stage 1 extracts pose-invariant geometry nodes and texture super-nodes from sparsely sampled frames. Stage 2 constructs a spatial dynamic hypergraph to fuse heterogeneous features, followed by adaptive attention pooling to obtain frame-level representations. Stage 3 employs a Bidirectional Mamba (Bi-Mamba) block to efficiently capture long-range temporal dependencies, concluding with temporal aggregation and a classification head for the final tri-class prediction.}
  \label{fig:pipeline}
\end{figure*}

Given a sparsely sampled long video sequence, the proposed framework is designed to accurately assess the global fatigue state by jointly capturing high-order spatial facial synergies and long-range temporal dynamics. An overview of our method is illustrated in Figure \ref{fig:pipeline}. We first detail the Global Sparse Sampling strategy, which extracts a sequence of $T$ frames from the raw video (Sec. 3.1). Following this, we introduce the Heterogeneous Feature Construction process, which extracts pose-disentangled geometric nodes and appearance-based texture super-nodes to form robust dual-stream representations (Sec. 3.2). Then, we present the core component of our spatial representation: a Spatial Dynamic Hypergraph Modeling module. This module fuses multimodal cues through a hierarchical Tex-Geo hypergraph and utilizes an adaptive attention pooling mechanism to obtain frame-level features (Sec. 3.3). Finally, we elaborate on the Temporal Evolution Modeling stage, where a Bidirectional Mamba (Bi-Mamba) block is employed to efficiently encapsulate both forward and backward temporal contexts with linear complexity for the final tri-class prediction (Sec. 3.4).

\subsection{Global Sparse Sampling}

Traditional methods often employ dense sliding windows, which not only incur massive computational overhead but also suffer from restricted local receptive fields, failing to capture the global context of a prolonged behavior. Therefore, we introduce a global sparse sampling strategy.

Given an untrimmed raw video $V$ consisting of $N$ frames, we uniformly sample a fixed-length sequence to represent the entire video dynamically. Mathematically, the raw video consists of several frames $S_{raw} = \{s_1, s_2, ..., s_N\}$, the sampled frame sequence $S = \{s_1, s_2, ..., s_T\}$ is obtained by selecting frames at specific indices $i_t \in \{1, 2, ..., N\}$, formulated as:
\begin{equation}
    i_t=\lfloor\frac{N}{T}\cdot t\rfloor,\quad t\in\{1,2,...,T\}
\end{equation}
where $T$ is the predefined sequence length. In our implementation, we set $T=128$. This sparse paradigm reduces the spatial redundancy of adjacent frames while granting the subsequent Bi-Mamba module a global temporal receptive field covering the entire video sequence.
%%%% 已改{end}
\subsection{Heterogeneous Feature Construction}
%%%% 已改{begin}
Upon obtaining the sparsely sampled frame sequence $S$, we extract dual-stream representations—geometry and texture—to capture complementary facial cues.
\\ \textbf{Geometry Stream and 3D Canonical Alignment.} Facial landmarks provide highly compressed geometric topology. Although tools like MediaPipe\cite{lugaresi2019mediapipeframeworkbuildingperception} excel at predicting dense 3D facial meshes in real-time, directly constructing a graph with hundreds of nodes introduces severe computational redundancy. Therefore, we strategically sample a semantic subset of 68 keypoints from the MediaPipe mesh, rigorously aligned with the standard Dlib 68-point protocol\cite{10.5555/1577069.1755843}. This mapping preserves crucial structural semantics while maintaining a lightweight node dimension.

Let $P_t \in \mathbb{R}^{68 \times 3}$ denote the raw 3D coordinates of the 68 landmarks at frame $s_t$. In driving scenarios, $P_t$ is highly entangled with rigid head movements, which can easily overwhelm the subtle non-rigid deformations caused by fatigue. Therefore, we introduce a 3D canonical alignment mechanism. Given a pre-defined frontal canonical face template $P_{ref} \in \mathbb{R}^{68 \times 3}$, we estimate the optimal scale $c_t$, rotation matrix $R_t \in \mathbb{R}^{3 \times 3}$, and translation vector $t_t \in \mathbb{R}^{3}$ by minimizing the Procrustes distance:
\begin{equation}
    \arg\min_{c_t,R_t,t_t}||c_tP_tR_t+t_t-P_{ref}||_F^2\quad\mathrm{s.t.}\quad R_t^TR_t=I
\end{equation}

By resolving this optimization, we transform the raw landmarks into a pose-invariant canonical space. The aligned coordinates, denoted as $X_t^{geo} \in \mathbb{R}^{68 \times 3}$, exclusively reflect pure facial muscular deformations. These canonical coordinates $X_t^{geo}$ serve as the initial features for the geometric nodes in our subsequent spatial hypergraph.%%%已改{end} 已改{begin}
\\ \textbf{Texture Stream and Micro-CNN Encoder.} While geometric landmarks can capture structural deformations, they intrinsically lack the visual semantics to distinguish specific physiological states, such as the subtle textural difference between an open and a closed eye. Therefore, a parallel texture stream is introduced. We select three localized Regions of Interest (ROIs)—corresponding to the left eye, right eye, and mouth—from the raw RGB frames. To guarantee computational efficiency, each ROI is resized to a compact resolution of $32 \times 32$ pixels.

Subsequently, these patches are fed into a lightweight Micro-CNN encoder and projected into a high-dimensional feature space independently. The resulting sequence of embeddings establishes three distinct "Texture Super-Nodes", denoted as $X_t^{tex}$, which reflects facial appearance states such as eye closure degrees. Finally, the geometry nodes $X_t^{geo}$ and the texture super-nodes $X_t^{tex}$ are concatenated across the feature dimension, formulating a hierarchical multimodal node set.

\subsection{Spatial Dynamic Hypergraph Modeling}

To ensure semantic alignment across modalities, the raw 3D coordinates of the geometry nodes $X_t^{geo}$ are first mapped into a shared $d$-dimensional latent space via a linear projection layer. Aiming to model the high-order synergistic deformations of facial muscles, we formulate the hierarchical nodes into a heterogeneous hypergraph $\mathcal{G} = (\mathcal{V}, \mathcal{E}, \mathbf{W})$. Let $\mathcal{V} = \mathcal{V}_{geo} \cup \mathcal{V}_{tex}$ denote the multimodal node set containing $N_{total} = 71$ vertices (i.e., 68 geometric nodes and 3 texture super-nodes). Their concatenated feature matrix is denoted as $\mathbf{X} = [\mathbf{X}^{geo} \parallel \mathbf{X}^{tex}] \in \mathbb{R}^{N_{total} \times d}$.

The core of our spatial modeling is constructing a heterogeneous incidence matrix $\mathbf{H} \in \mathbb{R}^{N_{total} \times E}$ that represents the topological connections. Unlike standard graphs restricted to pairwise edges, a hyperedge $e \in \mathcal{E}$ can connect an arbitrary number of vertices. Our incidence matrix consists of two complementary sub-matrices: $\mathbf{H} = [\mathbf{H}_{geo} \parallel \mathbf{H}_{tex}]$.

$\mathbf{H}_{geo}$ represents Geo-Geo Hyperedges, which aims to capture local geometric structures. We compute pairwise Euclidean distances based on the canonical coordinates $X_t^{geo}$. For each node, a hyperedge is constructed by connecting it to its $k$-nearest neighbors, forming $E_{geo} = 68$ spatial hyperedges.

$\mathbf{H}_{tex}$ denotes Tex-Geo Hyperedges. To achieve cross-modal fusion, we establish a star-topology broadcasting mechanism and define $E_{tex} = 3$ hyperedges corresponding to the left eye, right eye, and mouth regions. The incidence value $h(v, e)$ is formally defined as:
\begin{equation}
    h(v,e)=
    \begin{cases}
    1, & v\in\mathcal{R}_e \\
    0, & v\notin\mathcal{R}_e 
    \end{cases}
\end{equation}
where $\mathcal{R}_e \subset \mathcal{V}_{geo}$ denotes the predefined subset of geometric nodes associated with the localized related region of super-node $e$.

By cascading $\mathbf{H}_{geo}$ and $\mathbf{H}_{tex}$, the resulting incidence matrix $\mathbf{H}$ bridges localized visual semantics with global structural dynamics without triggering feature dimension explosion.

With the incidence matrix $\mathbf{H}$ established, we propagate the heterogeneous node features to capture high-order intra-modal and cross-modal interactions. Given the input node features $\mathbf{X}$, the hypergraph convolution layer updates the representations through a symmetric normalized Laplacian propagation\cite{Feng_You_Zhang_Ji_Gao_2019}:
\begin{equation}
    \mathbf{X}_{out}=\sigma\left(\mathbf{D}_v^{-1/2}\mathbf{H}\mathbf{W}\mathbf{H}^T\mathbf{D}_v^{-1/2}\mathbf{X}\mathbf{\Theta}\right)
\end{equation}
where $\mathbf{\Theta} \in \mathbb{R}^{d \times d_{out}}$ is the learnable weight matrix for linear feature transformation. $\mathbf{W} \in \mathbb{R}^{E \times E}$ is a learnable diagonal matrix representing the weight of each hyperedge, enabling the network to dynamically emphasize crucial spatial synergies. We define the weighted adjacency matrix as $\mathbf{A} = \mathbf{H} \mathbf{W} \mathbf{H}^T$. Consequently, $\mathbf{D}_v$ is the diagonal node degree matrix where its diagonal element is computed as the row sum of $\mathbf{A}$ (i.e., $\mathbf{D}_v(i,i) = \sum_{j} \mathbf{A}_{i,j}$). Finally, $\sigma(\cdot)$ denotes the LeakyReLU activation function. Through this layer, geometric landmarks can obtain collaborative information from each other and visual semantics from texture super nodes. 

Following the hypergraph convolution, the geometric nodes $\mathbf{X}^{geo}_{out} \in \mathbb{R}^{N_{geo} \times d_{out}}$ are highly enriched with multimodal context. Since the texture super-nodes have fulfilled their broadcasting mission, we safely discard them to focus exclusively on the facial topology. To compress these node-level features into a compact, frame-level representation, we introduce an Adaptive Attention Pooling module.

We deploy a Multi-Layer Perceptron with a $\tanh$ activation to estimate the distinct contribution of each node dynamically. The attention score $e_i$ and the normalized weight $\alpha_i$ for the $i$-th node are calculated as:
\begin{equation}e_i=\mathbf{w}_a^T\tanh(\mathbf{W}_a\mathbf{x}_i+\mathbf{b}_a)\end{equation}
\begin{equation}\alpha_i=\frac{\exp(e_i)}{\sum_{j=1}^{N_{geo}}\exp(e_j)}\end{equation}
where $\mathbf{W}_a, \mathbf{b}_a,$ and $\mathbf{w}_a$ are learnable parameters of the attention network. Finally, the frame-level feature vector $\mathbf{z}_{seq} \in \mathbb{R}^{d_{out}}$ is obtained via a weighted sum: $\mathbf{z}_{seq} = \sum_{i=1}^{N_{geo}} \alpha_i \mathbf{x}_i$. This mechanism inherently filters out irrelevant localized noise and selectively forces the model to attend to salient fatigue indicators, formulating a robust temporal sequence for the subsequent Bi-Mamba module.
%已改{end}

\subsection{Temporal Evolution Modeling}

After spatial modeling, the untrimmed video is abstracted into a sequence of highly refined frame-level features $\mathbf{Z} = [\mathbf{z}_1, \mathbf{z}_2, \dots, \mathbf{z}_T] \in \mathbb{R}^{T \times d_{out}}$. To capture the long-range temporal dependencies of fatigue behaviors, traditional RNNs suffer from long-term forgetting, while Transformers incur an heavy $O(T^2)$ computational complexity. To resolve this dilemma, we introduce a Temporal Evolution Modeling module driven by a Bidirectional State Space Model (Bi-Mamba).

The core of Mamba originates from the continuous-time State Space Model (SSM)\cite{Gu2021EfficientlyML}, which maps a 1D input sequence $z(t)$ to an output response $y(t)$ via a latent state $h(t)$. Mathematically, it is formulated as a linear Ordinary Differential Equation:
\begin{equation}
    h^{\prime}(t)=\mathbf{A}h(t)+\mathbf{B}z(t),\quad y(t)=\mathbf{C}h(t)
\end{equation}
where $\mathbf{A}$ acts as the evolution matrix, and $\mathbf{B}, \mathbf{C}$ are projection parameters. To process discrete video frames, we apply the zero-order hold rule with a timescale parameter $\Delta$ to discretize the continuous matrices into $\mathbf{\bar{A}}$ and $\mathbf{\bar{B}}$:
\begin{equation}
    \bar{\mathbf{A}}=\exp(\Delta\mathbf{A}),\quad\bar{\mathbf{B}}=(\Delta\mathbf{A})^{-1}(\exp(\Delta\mathbf{A})-\mathbf{I})\cdot\Delta\mathbf{B}
\end{equation}

Crucially, standard SSMs utilize time-invariant parameters. In contrast, our module leverages the selective scan mechanism\cite{gu2024mambalineartimesequencemodeling}, parameterizing $\mathbf{B}, \mathbf{C},$ and $\Delta$ as data-dependent functions of the input $\mathbf{z}_t$. This input-awareness empowers the model to dynamically filter out redundant information (e.g., prolonged periods of normal driving) and memorize salient occurrences (e.g., the onset of a yawn) into the hidden state.

Standard mamba has a causal, unidirectional nature—it processes sequences strictly forward. However, assessing complex behaviors often requires offline temporal localization where future context is necessary for understanding current actions. Therefore, we deploy a Bi-Mamba block comprising a forward SSM and a backward SSM. Given the input sequence, the forward scan processes it chronologically to yield $\overrightarrow{\mathbf{Y}}$, while the backward scan processes the reversed sequence to yield $\overleftarrow{\mathbf{Y}}$. The bidirectional temporal representation is then obtained via feature summation:
\begin{equation}
    \mathbf{Y}_{bi}=\overrightarrow{\mathbf{Y}}+\overleftarrow{\mathbf{Y}}\in\mathbb{R}^{T\times d_{out}}
\end{equation}

Finally, since fatigue events may occur sparsely within the 128-frame sequence, mean pooling might dilute these critical signals. Thus, we perform a global temporal max pooling across the temporal dimension to extract the most discriminative global video vector $\mathbf{v}_{final}$:
\begin{equation}
    \mathbf{v}_{final}=\max_{t=1}^T(\mathbf{Y}_{bi}^{(t)})
\end{equation}

This vector is subsequently fed into a fully connected classification head with a Softmax activation to predict the ultimate behavior category (Normal, Talking, or Yawning).

\subsection{Optimization and Loss Function}

For the reason of two inherent challenges in unconstrained driving datasets: severe class imbalance and subtle intra-class variations, We tackle this by jointly optimizing a multi-class Focal Loss $\mathcal{L}_{foc}$ and a Center Loss $\mathcal{L}_{cen}$.

To counteract class imbalance and heavily penalize hard-to-distinguish boundary samples such as an onset of yawning versus talking, $\mathcal{L}_{foc}$ is formulated as:
\begin{equation}\mathcal{L}_{foc}=-\sum_{c=1}^C\alpha_c(1-p_c)^\gamma y_c\log(p_c)\end{equation}
where $C=3$, $y_c$ is the one-hot label, and the modulating factor $(1 - p_c)^\gamma$ dynamically scales the loss based on prediction confidence.

Simultaneously, to enforce intra-class compactness regardless of individual identity or head pose, $\mathcal{L}_{cen}$ minimizes the distance between the temporal feature $\mathbf{v}_{final}$ and its corresponding learnable class center $\mathbf{c}_{y_i}$:
\begin{equation}\mathcal{L}_{cen}=\frac{1}{2}\sum_{i=1}^B||\mathbf{v}_{final}^{(i)}-\mathbf{c}_{y_i}||_2^2\end{equation}
where $B$ is the batch size. The overall objective function is optimized as $\mathcal{L}_{total} = \mathcal{L}_{foc} + \lambda \mathcal{L}_{cen}$, where the hyperparameter $\lambda$ balances the two terms.
%% 已改{end}
\section{Experiments}
\subsection{Datasets and Implementation Details}
\textbf{Dataset.} To evaluate the robustness and generalizability of our proposed HST-HGN framework, we conduct experiments on a diverse collection of real-world datasets. Our primary training and evaluation benchmark is derived from the widely adopted YawDD dataset\cite{e1qm-hb90-20}. To construct a high-fidelity benchmark, we split and relabel sequences containing overlapping talking and yawning instances. Then, we recursively truncate long sequences (duration > 30s) and discard overly short segments. This curation yields a clean, non-overlapping corpus of 427 pure video clips. To fundamentally prevent data leakage and assess true cross-identity generalization, we strictly enforce a subject-independent (driver-level) split. The dataset is partitioned into training (70\%), validation (15\%), and testing (15\%) sets based on unique subject IDs, intrinsically preserving the real-world class imbalance. 

To further establish a comprehensive cross-domain generalization analysis, our evaluation uses four additional fatigue datasets: UTA-RLDD\cite{ghoddoosian2019realisticdatasetbaselinetemporal}, FatigueView\cite{Yang2022FatigueView}, and DMD\cite{10.1007/978-3-030-66823-5_23}. These datasets introduce varied camera perspectives, diverse illumination conditions, and heterogeneous behavioral annotations. Due to the missing Talking labels and samples in some datasets, all sequence annotations are roughly mapped into a binary classification of Yawning and Normal.
\\ \textbf{Implementation Details.} All experiments are conducted using PyTorch(v2.4.0) and accelerated via CUDA 3.11 on a system equipped with a single NVIDIA RTX 3060 GPU. This hardware constraint deliberately underscores the lightweight and edge-deployment feasibility of our architecture. For spatial modeling, the spatial hypergraph is constructed dynamically with $K=4$ nearest neighbors. The entire network is trained from scratch for 100 epochs with a batch size of 8.

The main network parameters are optimized using the Adam optimizer with an initial learning rate of $1 \times 10^{-3}$ and a weight decay of $1 \times 10^{-4}$ to prevent overfitting. Concurrently, the learnable class centers formulated in the Center Loss are updated using Stochastic Gradient Descent (SGD) with a significantly larger learning rate of 0.5. For the loss formulation, the Focal Loss parameters are set to $\alpha=0.25$ and $\gamma=2.0$ to heavily penalize hard-to-distinguish boundary samples, while the Center Loss weight is set to $\lambda=0.001$ to enforce intra-class compactness.

\subsection{Comparison with SOTA Methods}
\begin{table}[t]
\centering
\caption{Comparison with state-of-the-art generic video models and recent domain-specific fatigue detection networks on the YawDD dataset. The best and second-best results are highlighted in bold and underlined, respectively.}
\label{tab:yawdd_sota}
\resizebox{\linewidth}{!}{
\begin{tabular}{l l c c}
\toprule
\textbf{Category} & \textbf{Method} & \textbf{Acc (\%)} & \textbf{Macro F1 (\%)} \\
\midrule
\multirow{2}{*}{\textit{Generic Models}} 
 & SlowFast  & 92.86 & 91.22 \\
 & VideoMAE  & 92.86 & 91.59 \\
\midrule
\multirow{5}{*}{\textit{Fatigue-Specific}} 
 & 2s ST-GCN  & 90.00 & 89.37 \\
 & VBFLLFA  & 91.43 & 89.01 \\
 & JHPFA-Net  & 94.29 & 93.18 \\
 & IsoSSL-MoCo  & 95.71 & 95.08 \\
 & LiteFat  & \underline{97.14} & \underline{96.59} \\
\midrule
\textbf{Ours} & \textbf{HST-HGN} & \textbf{98.57} & \textbf{98.28} \\
\bottomrule
\end{tabular}
}
\end{table}
We compare our proposed HST-HGN with a comprehensive suite of state-of-the-art (SOTA) methods on the YawDD dataset. To ensure a rigorous evaluation, our baselines include foundational generic video understanding architectures (SlowFast\cite{9008780} and VideoMAE\cite{NEURIPS2022_416f9cb3}) and the most recent, domain-specific driver fatigue detection models. Specifically, we benchmark against Lightweight Spatial-Temporal Graph Learning (LiteFat)\cite{11247563}, Joint Head Pose and Facial Action Network (JHPFA-Net)\cite{10159554}, Isotropic Self-Supervised Learning with Attention-Based Multimodal Fusion (IsoSSL-MoCo)\cite{9618813}, Video-Based Driver Drowsiness Detection With Optimised Utilization of Key Facial Features (VBFLLFA)\cite{10382460}, and 2s ST-GCN\cite{9557757}.

As reported in Table \ref{tab:yawdd_sota}, generic models like VideoMAE achieve robust accuracy (92.86\%). Moreover, recent specialized models like LiteFat show strong performance (97.14\%) by focusing on specific facial actions. Nevertheless, our HST-HGN outperforms all competitors, achieving the Top-1 Accuracy of 98.57\% and a Macro F1-Score of 98.28\%. By integrating lightweight Micro-CNN texture features with pose-disentangled geometric landmarks via dynamic hypergraph convolutions, HST-HGN extracts highly discriminative representations, proving exceptionally effective at identifying driving fatigue.

To further demonstrate the transferability of HST-HGN, we conduct cross-dataset evaluation on UTA-RLDD, FatigueView, and DMD using the model trained on YawDD. As shown in Table \ref{tab:cross_dataset}, HST-HGN consistently maintains robust predictive performance across domain shifts. Compared to existing specific fatigue models, HST-HGN establishes a new state-of-the-art for domain-generalized fatigue assessment.
\begin{table}[t] 
\centering
\caption{Cross-dataset generalization performance (Binary classification). The classification head is adapted while the backbone remains unchanged to evaluate domain transferability.}
\label{tab:cross_dataset}
\resizebox{\linewidth}{!}{
\begin{tabular}{l | c c | c c | c c}
\toprule
\multirow{2}{*}{\textbf{Method}} & \multicolumn{2}{c|}{\textbf{UTA-RLDD}} & \multicolumn{2}{c|}{\textbf{FatigueView}} & \multicolumn{2}{c}{\textbf{DMD}} \\
\cline{2-7}
 & Acc (\%) & F1 (\%) & Acc (\%) & F1 (\%) & Acc (\%) & F1 (\%) \\
\midrule
SlowFast  & 87.32 & 85.91 & 90.06 & 89.73 & 92.75 & 92.60 \\
VideoMAE  & 88.03 & 87.61 & 89.44 & 87.87 & 90.13 & 90.98 \\
\midrule
2s ST-GCN  & 81.34 & 80.86 & 87.58 & 86.80 & 90.82 & 90.34 \\
VBFLLFA  & 88.73 & 88.40 & 90.68 & 90.43 & 90.82 & 90.80 \\
JHPFA-Net  & 92.25 & \underline{91.94} & 93.17 & 92.81 & 92.27 & 91.99 \\
IsoSSL-MoCo  & 90.49 & 89.74 & 92.55 & 92.22 & \underline{96.62} & \underline{96.52} \\
LiteFat  & \underline{92.61} & 91.58 & \underline{93.79} & \underline{93.26} & 96.14 & 96.05 \\
\midrule
\textbf{HST-HGN (Ours)} & \textbf{94.72} & \textbf{94.34} & \textbf{94.41} & \textbf{94.08} & \textbf{97.10} & \textbf{96.99} \\
\bottomrule
\end{tabular}
}
\end{table}

\subsection{Efficiency Analysis}
%=================================================
% Table 4: Efficiency comparison on a 128-frame sequence.
%=================================================
\begin{table}[t]
\centering
\caption{Efficiency comparison on a 128-frame sequence. All profiling metrics are measured with a batch size of 1 on a single NVIDIA RTX 3060. The best and second-best results are bolded and underlined respectively.}
\label{tab:efficiency}
\resizebox{\linewidth}{!}{
\begin{tabular}{l c c c c c c}
\toprule
\textbf{Method} & \textbf{Params} & \textbf{FLOPs} & \textbf{MACs} & \textbf{VRAM} & \textbf{Latency} & \textbf{Throughput} \\
 & (M) & (G) & (G) & (MB) & (ms) & (Clips/s) \\
\midrule
SlowFast & 33.65 & 406.84 & 203.42 & 702.98 & 103.31 & 9.68 \\
VideoMAE & 64.98 & 1629.58 & 814.79 & 1002.06 & 2564.10 & 0.39 \\
IsoSSL-MoCo & 33.88 & 455.93 & 227.97 & 672.84 & 67.61 & 14.79 \\
JHPFA-Net & 7.77 & 106.83 & 53.42 & 1404.55 & 99.50 & 10.05 \\
2s ST-GCN & 3.07 & 18.95 & 9.48 & \textbf{55.46} & \textbf{7.44} & \textbf{134.37} \\
LiteFat & \underline{1.32} & \underline{17.25} & \underline{8.62} & 405.90 & 59.77 & 16.73 \\
\midrule
\textbf{HST-HGN (Ours)} & \textbf{0.30} & \textbf{2.90} & \textbf{1.45} & \underline{71.91} & \underline{37.81} & \underline{26.45} \\
\bottomrule
\end{tabular}
}
\end{table}
Beyond predictive accuracy, we evaluate the real-time deployment potential of HST-HGN (Table \ref{tab:efficiency}). Generic heavyweight models, such as VideoMAE and SlowFast, impose massive computational burdens exceeding 400 G FLOPs and 30 M parameters, making them fundamentally impractical for resource-constrained in-cabin edge devices. Even when compared to domain-specific lightweight architectures like LiteFat and JHPFA-Net, our proposed HST-HGN demonstrates superiority in resource optimization. Specifically, it comprises merely 299 K trainable parameters and requires an ultra-low computation of 2.90 G FLOPs (1.45 G MACs) per 128-frame clip. 

Although the inevitable memory I/O overhead from dynamic multi-modal texture cropping yields a comparatively lower throughput (26.45 Clips/s) than pure skeleton-based networks like 2s ST-GCN, this design guarantees a dominant 98.57\% accuracy. By satisfying the standard 25 FPS real-time requirement, HST-HGN achieves a trade-off between hardware-level efficiency and discriminative power.

\subsection{Ablation Studies}
%====================================================================
% Table 3: Detailed Ablation Studies on the YawDD Dataset
%====================================================================
\begin{table*}[t]
\centering
\caption{Detailed ablation studies validating the contribution of each component in the proposed HST-HGN framework. The evaluation metric tracks the progression from a naive baseline to the final sophisticated architecture.}
\label{tab:ablation_detailed}
\scalebox{0.95}{
\begin{tabular}{l l c l c l c c c}
\toprule
\textbf{No.} & \textbf{Variant} & \textbf{3D Align} & \textbf{Spatial} & \textbf{Texture} & \textbf{Temporal} & \textbf{Loss} & \textbf{Acc (\%)} & \textbf{Macro F1 (\%)} \\
\midrule
\#1 & Baseline & - & GCN & - & MaxPool & CE & 80.00 & 80.78 \\
\#2 & +3D Align & \checkmark & GCN & - & MaxPool & CE & 84.29 & 85.01 \\
\#3 & +HGNN & \checkmark & HGNN & - & MaxPool & CE & 90.00 & 89.27 \\
\#4 & +Texture & \checkmark & HGNN & \checkmark & MaxPool & CE & 92.86 & 91.90 \\
\#5 & +Bi-Mamba & \checkmark & HGNN & \checkmark & Bi-Mamba & CE & 95.71 & 95.22 \\
\midrule
\textbf{\#6} & \textbf{HST-HGN} & \textbf{\checkmark} & \textbf{HGNN} & \textbf{\checkmark} & \textbf{Bi-Mamba} & \textbf{Focal + Center} & \textbf{98.57} & \textbf{98.28} \\
\bottomrule
\end{tabular}
}
\end{table*}
\textbf{Effectiveness of Proposed Components.} We establish our baseline (Model \#1) as a naive spatial-temporal network utilizing raw unaligned facial landmarks, a standard Graph Convolutional Network (GCN) for spatial modeling, simple Max Pooling for temporal aggregation, and optimized via standard Cross-Entropy (CE) loss. The incremental performance gains achieved by integrating our core modules are detailed in Table \ref{tab:ablation_detailed}.

The baseline model (\#1) struggles to discern dynamic fatigue shifts, yielding only 80.00\% accuracy. Introducing 3D alignment (\#2) effectively mitigates head pose variations (+4.29\%). Building upon this, replacing the GCN with our proposed HGNN (\#3) leads to a leap to 90.00\%. Furthermore, injecting local appearance cues via multimodal texture fusion (\#4) pushes accuracy to 92.86\%, proving that geometric topology alone is insufficient for subtle micro-expressions. Temporally, rather than naive pooling, the Bi-Mamba module (\#5) dynamically captures long-range dependencies, driving the accuracy to 95.71\% and F1-Score to 95.22\%. Finally, joint optimization with Focal and Center Loss (\#6) resolves the long-tail class imbalance and compacts intra-class variations, achieving the peak state-of-the-art performance of 98.57\% accuracy and 98.28\% F1-Score.
\begin{table*}[t]
\centering
\caption{Comparison of different temporal modeling architectures. All models are evaluated with a batch size of 8. The best and second-best performance metrics are highlighted in \textbf{bold} and underlined, respectively.}
\label{tab:temporal_ablation}
\resizebox{\textwidth}{!}{
\begin{tabular}{l l c c c c c c c c}
\toprule
\multirow{2}{*}{\textbf{Category}} & \multirow{2}{*}{\textbf{Method}} & \textbf{Params} & \textbf{MACs} & \textbf{FLOPs} & \textbf{VRAM} & \textbf{Latency} & \textbf{Throughput} & \textbf{Acc} & \textbf{Macro F1} \\
 & & (M) & (G) & (G) & (MB) & (ms) & (Clips/s) & (\%) & (\%) \\
\midrule
\multirow{5}{*}{\textit{RNN \& CNN-based}} 
 & RNN & \underline{0.108} & \underline{1.424} & \underline{2.849} & \textbf{470.96} & \textbf{40.84} & \textbf{195.89} & 88.57 & 88.33 \\
 & LSTM & 0.207 & 1.437 & 2.874 & 471.34 & 41.58 & 192.40 & 87.14 & 86.56 \\
 & GRU & 0.174 & 1.433 & 2.866 & 471.22 & \underline{40.98} & \underline{195.22} & 91.43 & 89.98 \\
 & BiLSTM & 0.364 & 1.455 & 2.830 & 471.22 & 41.22 & 194.08 & 90.00 & 89.34 \\
 & TCN & 0.173 & 1.433 & 2.865 & \underline{471.21} & 41.00 & 195.12 & 92.86 & 91.97 \\
\midrule
\multirow{4}{*}{\textit{Attention-based}} 
 & Transformer & 0.207 & 1.438 & 2.876 & 472.02 & 41.28 & 193.80 & 85.71 & 87.14 \\
 & Informer\cite{zhou2021informer} & 0.205 & 1.438 & 2.875 & 472.02 & 41.44 & 193.05 & 94.29 & 94.09 \\
 & Autoformer\cite{NEURIPS2021_bcc0d400} & 0.158 & 1.431 & 2.861 & 472.63 & 41.28 & 193.80 & 87.14 & 86.85 \\
 & RetNet\cite{sun2023retentivenetworksuccessortransformer} & 0.215 & 1.437 & 2.876 & 472.82 & 41.36 & 193.42 & 81.43 & 81.87 \\
\midrule
\multirow{4}{*}{\textit{SSM-based}} 
 & S4 & \textbf{0.094} & \textbf{1.422} & \textbf{2.845} & 473.18 & \underline{40.98} & \underline{195.22} & 95.71 & 95.28 \\
 & Mamba & 0.191 & 1.434 & 2.869 & 473.56 & 41.12 & 194.55 & 91.43 & 91.11 \\
 & VMamba\cite{NEURIPS2024_baa2da9a} & 0.341 & 1.453 & 2.906 & 471.85 & 41.68 & 191.94 & \underline{97.14} & \underline{96.99} \\
 & \textbf{BiMamba (Ours)} & 0.308 & 1.449 & 2.897 & 473.01 & 41.60 & 192.31 & \textbf{98.57} & \textbf{98.28} \\
\bottomrule
\end{tabular}
}
\end{table*}
\noindent \textbf{Effectiveness of the Temporal Modeling Architecture.} To validate the superiority of our proposed BiMamba module, we conducted an ablation study by replacing it with various mainstream temporal modeling architectures. As detailed in Table \ref{tab:temporal_ablation}, the baselines are grouped into three distinct categories: RNN \& CNN-based, Attention-based, and SSM-based models. Since the spatial hypergraph backbone remains frozen during this evaluation, all variants exhibit highly consistent results.

However, significant disparities emerge in predictive performance. Traditional RNNs and CNNs, alongside standard Attention-based models such as Transformer and Informer, struggle to fully capture the complex, long-range temporal dynamics of fatigue behaviors, yielding sub-optimal Macro F1 scores. In contrast, SSM-based architectures demonstrate a clear advantage in sequence reasoning. While recent state-of-the-art SSMs like S4 and VMamba show strong potential, our BiMamba explicitly captures bidirectional contextual dependencies. This design enables the network to comprehensively model the complete physiological lifecycle of transient actions—such as the onset, apex, and offset of a yawn or blink. Consequently, BiMamba achieves a dominant peak Accuracy of 98.57\% and a Macro F1 of 98.28\% without introducing noticeable computational burdens, solidifying its role as the optimal temporal engine for our framework.
\begin{figure}[t]
    \centering
    \includegraphics[width=\linewidth]{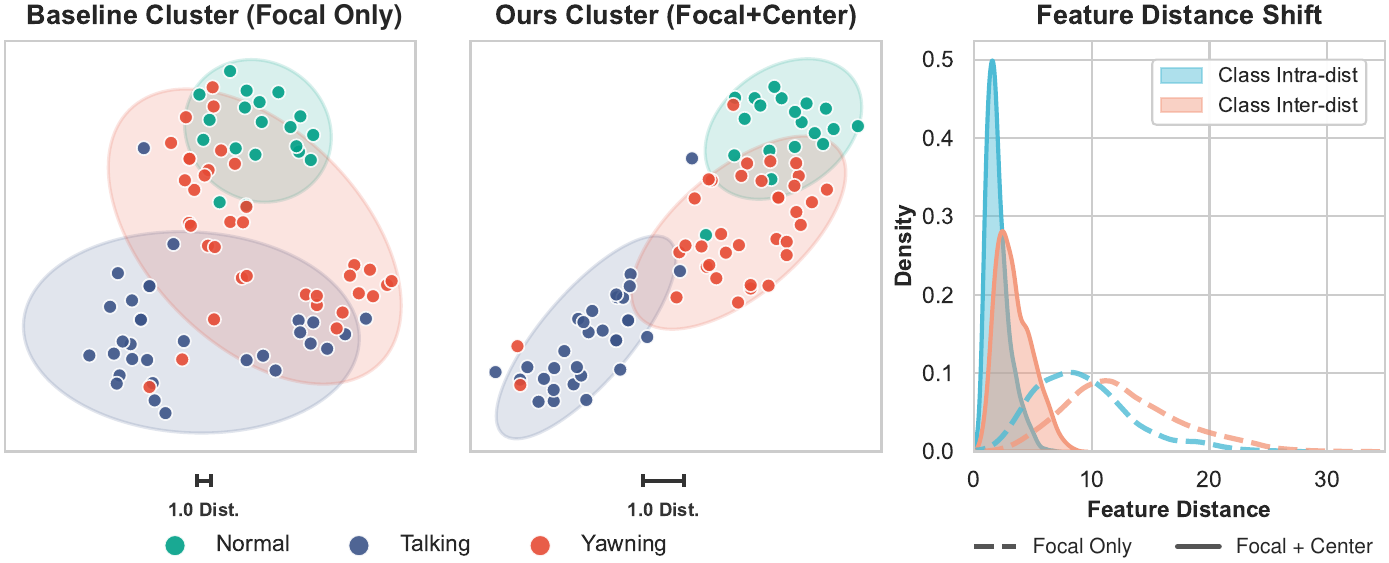} 
    \caption{Qualitative \textbf{t-SNE} visualization of the learned feature space. \textbf{Left:} The baseline clusters of only Focal Loss. \textbf{Middle:}Well-separated clusters of our joint optimization combining both Focal Loss and Center Loss. \textbf{Right:} The feature distance shift proves that the Center Loss (solid lines) compresses intra-class variance compared to using only Focal Loss (dashed lines).}
    \label{fig:feature_shift}
\end{figure}
\noindent \textbf{Effectiveness of Joint Loss Optimization.} To demonstrate the impact of our joint optimization strategy (Focal + Center Loss), we visualize the learned feature space using t-SNE in Figure \ref{fig:feature_shift}. As shown in the Baseline Cluster (left), relying solely on focal loss results in severe feature entanglement, which is adverse to distinguishing between the ambiguous "Talking" and "Yawning" classes. However, optimized with our joint loss, the middle figure exhibits highly cohesive and separable boundaries. This qualitative observation is mathematically validated by the Feature Distance Shift distributions (right). When using only Focal Loss, the intra-class and inter-class distances overlap heavily, leading to inevitable misclassifications. The introduction of Center Loss compresses the intra-class distance into a sharp peak near zero while maintaining a distinct margin from the inter-class distance. This explicitly proves that our framework effectively pulls intra-class samples toward their respective learned centers, thereby significantly enhancing the model's discriminative power against subtle facial behaviors.
\subsection{Qualitative Analysis and Interpretability}
\begin{figure}[htbp]
    \centering
    \includegraphics[width=\linewidth]{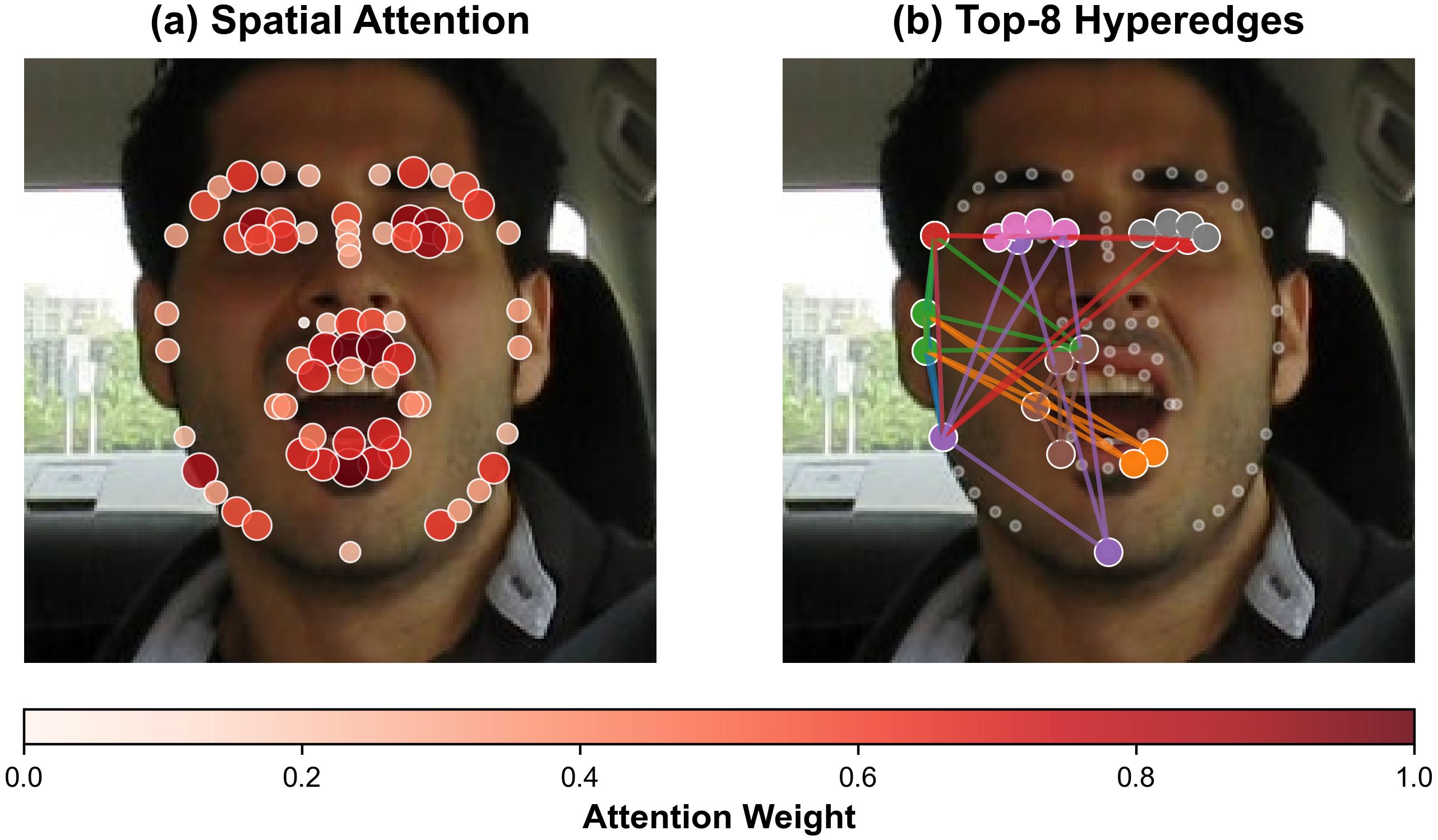}
    \caption{Visualization of spatial reasoning during a yawning event. (a) The spatial attention map highlights critical semantic regions (mouth and eyes) while suppressing irrelevant background noise. (b) The Top-8 active hyperedges illustrate how the proposed HGNN captures high-order, globally coordinated facial muscle deformations. Each color represents a hyperedge and the nodes connected by it.}
    \label{fig:spatial_attention}
\end{figure}

\noindent \textbf{Spatial Attention and Topology.} To elucidate the spatial reasoning mechanism, we visualize the learned attention weights and hypergraph topology during a yawning event (Figure \ref{fig:spatial_attention}). As shown in Figure \ref{fig:spatial_attention}(a), the node size and color intensity represent spatial attention weights. The model assigns the highest attention to the mouth, lips, and eye regions, which aligns perfectly with yawning physiological characteristics. This proves the network effectively suppresses irrelevant facial areas and focuses on discriminative semantic keypoints. Furthermore, Figure \ref{fig:spatial_attention}(b) illustrates the Top-8 most active hyperedges. Unlike standard pairwise graphs that only connect adjacent neighbors, our learned hyperedges span distinct facial regions (e.g., simultaneously linking the lower jaw, cheek, and eyes). This demonstrates the unique capability of HGNN to capture high-order, globally coordinated muscle deformations.
\begin{figure}[htbp]
    \centering
    \includegraphics[width=\linewidth]{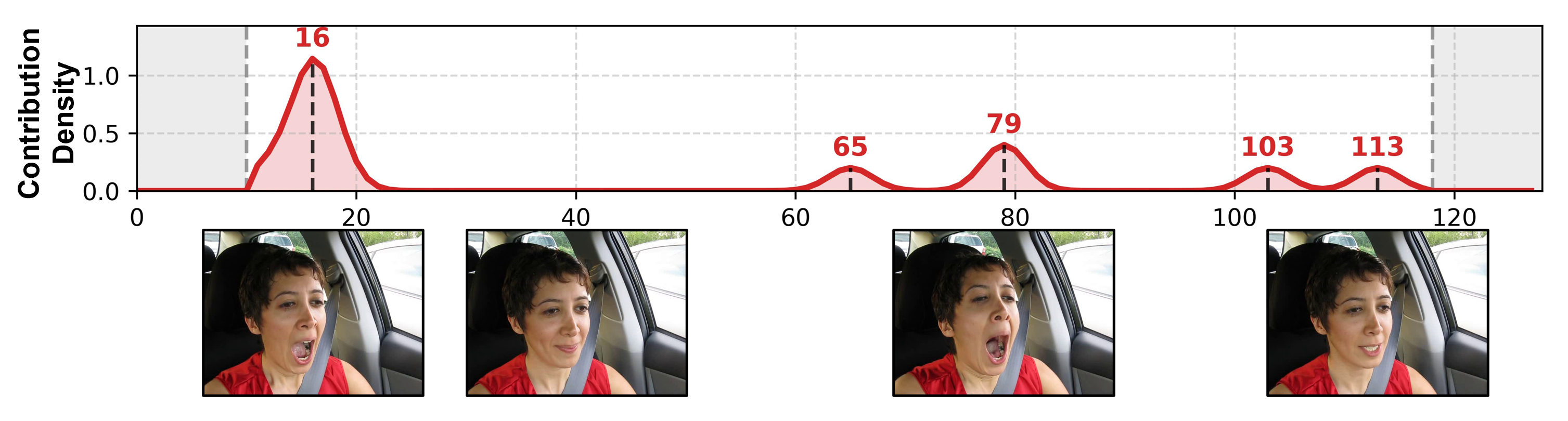}
    \caption{Visualization of the temporal contribution density over a 128-frame sequence. By tracking the indices of maximum activations during Global Max Pooling, the waveform reveals how the model explicitly assigns high decision weights to semantically critical frames while ignoring redundant background frames.}
    \label{fig:temporal_saliency}
\end{figure}

\noindent \textbf{Temporal Saliency and Reasoning.} To understand the internal temporal reasoning of HST-HGN, we visualize the temporal contribution density over a 128-frame input sequence. Since our framework employs Global Max Pooling to aggregate temporal features into a clip-level representation, we track the temporal indices of the maximum values across all feature channels. The resulting density waveform (Figure \ref{fig:temporal_saliency}) directly reflects the importance of each frame to the final classification decision. The model intelligently suppresses redundant normal driving frames (near-zero flat regions) and exhibits sharp response peaks precisely at frames containing critical fatigue semantics, demonstrating exceptional interpretability.
\section{Conclusion}
In this paper, we propose HST-HGN, a highly efficient framework tailored for real-time driver fatigue detection. By integrating spatial geometric topologies with multi-modal texture patches through a Hypergraph Neural Network, our method captures high-order facial muscle deformations, overcoming the limitations of traditional pairwise graphs. Furthermore, the BiMamba temporal module enables precise bidirectional modeling to explicitly filter noise and encompass the complete physiological lifecycle of transient fatigue actions. Extensive evaluations across diverse benchmarks demonstrate that HST-HGN achieves state-of-the-art performance. By striking an optimal balance between discriminative power and computational complexity, the proposed model ensures robust real-time throughput, making it well-suited for embedded edge applications. Future work will investigate extending our framework to handle extreme real-world scenarios, thereby further improving detection reliability in complex driving environments.
\newpage
\bibliographystyle{ACM-Reference-Format}
\bibliography{sample-sigconf-authordraft}

@ARTICLE{10517310,
  author={Fu, Biying and Boutros, Fadi and Lin, Chin-Teng and Damer, Naser},
  journal={IEEE Transactions on Intelligent Vehicles}, 
  title={A Survey on Drowsiness Detection: Modern Applications and Methods}, 
  year={2024},
  volume={9},
  number={11},
  pages={7279-7300}
}

@article{elnabi2024machine,
  author = {Abd El-Nabi, Samy and El-Shafai, Walid and El-Rabaie, El-Sayed M. and Ramadan, Khalil F. and Abd El-Samie, Fathi E. and Mohsen, Saeed},
  title = {Machine learning and deep learning techniques for driver fatigue and drowsiness detection: a review},
  journal = {Multimedia Tools and Applications},
  volume = {83},
  number = {3},
  pages = {9441--9477},
  year = {2024}
}

@ARTICLE{8482470,
  author={Sikander, Gulbadan and Anwar, Shahzad},
  journal={IEEE Transactions on Intelligent Transportation Systems}, 
  title={Driver Fatigue Detection Systems: A Review}, 
  year={2019},
  volume={20},
  number={6},
  pages={2339-2352}
}

@article{SUN2023106981,
title = {Facial feature fusion convolutional neural network for driver fatigue detection},
journal = {Engineering Applications of Artificial Intelligence},
volume = {126},
pages = {106981},
year = {2023},
issn = {0952-1976},


author = {Zhichao Sun and Yinan Miao and Jun Young Jeon and Yeseul Kong and Gyuhae Park}
}

@article{zhao2024review,
  author = {Zhao, Xia and Wang, Limin and Zhang, Yufei and Han, Xuming and Deveci, Muhammet and Parmar, Milan},
  title = {A review of convolutional neural networks in computer vision},
  journal = {Artificial Intelligence Review},
  volume = {57},
  pages = {99},
  year = {2024}
}

@article{FA2023103826,
title = {Multi-scale spatial–temporal attention graph convolutional networks for driver fatigue detection},
journal = {Journal of Visual Communication and Image Representation},
volume = {93},
pages = {103826},
year = {2023},
issn = {1047-3203},


author = {Shuxiang Fa and Xiaohui Yang and Shiyuan Han and Zhiquan Feng and Yuehui Chen}
}

@article{WU2025111106,
title = {Local and global self-attention enhanced graph convolutional network for skeleton-based action recognition},
journal = {Pattern Recognition},
volume = {159},
pages = {111106},
year = {2025},
issn = {0031-3203},


author = {Zhize Wu and Yue Ding and Long Wan and Teng Li and Fudong Nian}
}

@ARTICLE{10018105,
  author={Yang, Cuiliu and Pei, Zhao},
  journal={IEEE Transactions on Intelligent Transportation Systems}, 
  title={Long-Short Term Spatio-Temporal Aggregation for Trajectory Prediction}, 
  year={2023},
  volume={24},
  number={4},
  pages={4114-4126}
}

@article{ZHANG2024124124,
title = {A novel temporal adaptive fuzzy neural network for facial feature based fatigue assessment},
journal = {Expert Systems with Applications},
volume = {252},
pages = {124124},
year = {2024},
issn = {0957-4174},


author = {Zhimin Zhang and Hongmei Wang and Qian You and Liming Chen and Huansheng Ning}}

@article{hassan2025realtime,
  author = {Hassan, Osama F. and Ibrahim, Ahmed F. and Gomaa, Ahmed and Makhlouf, M. A. and Hafiz, Bassel},
  title = {Real-time driver drowsiness detection using transformer architectures: a novel deep learning approach},
  journal = {Scientific Reports},
  volume = {15},
  number = {1},
  pages = {17493},
  year = {2025}
}

@ARTICLE{10159554,
  author={Lu, Yansha and Liu, Chunsheng and Chang, Faliang and Liu, Hui and Huan, Hengqiang},
  journal={IEEE Transactions on Intelligent Transportation Systems}, 
  title={JHPFA-Net: Joint Head Pose and Facial Action Network for Driver Yawning Detection Across Arbitrary Poses in Videos}, 
  year={2023},
  volume={24},
  number={11},
  pages={11850-11863}
}

@inproceedings{li2024videomambastatespacemodel,
  author = {Kunchang Li and Xinhao Li and Yi Wang and Yinan He and Yali Wang and Limin Wang and Yu Qiao},
  title = {VideoMamba: State Space Model for Efficient Video Understanding},
  booktitle = {European Conference on Computer Vision},
  pages = {237--255},
  year = {2024},
  publisher = {Springer}
}

@INPROCEEDINGS{8953343,
  author={Wu, Chao-Yuan and Feichtenhofer, Christoph and Fan, Haoqi and He, Kaiming and Krähenbühl, Philipp and Girshick, Ross},
  booktitle={2019 IEEE/CVF Conference on Computer Vision and Pattern Recognition (CVPR)}, 
  title={Long-Term Feature Banks for Detailed Video Understanding}, 
  year={2019},
  volume={},
  number={},
  pages={284-293}
}

@article{SAHA2025130417,
title = {Vision transformers on the edge: A comprehensive survey of model compression and acceleration strategies},
journal = {Neurocomputing},
volume = {643},
pages = {130417},
year = {2025},
issn = {0925-2312},


author = {Shaibal Saha and Lanyu Xu}}

@InProceedings{pmlr-v139-bertasius21a,
  title = 	 {Is Space-Time Attention All You Need for Video Understanding?},
  author =       {Bertasius, Gedas and Wang, Heng and Torresani, Lorenzo},
  booktitle = 	 {Proceedings of the 38th International Conference on Machine Learning},
  pages = 	 {813--824},
  year = 	 {2021},
  editor = 	 {Meila, Marina and Zhang, Tong},
  volume = 	 {139},
  series = 	 {Proceedings of Machine Learning Research},
  month = 	 {18--24 Jul},
  publisher =    {PMLR}
}

@article{xxxxx,
author = {Zhao, Zuopeng and Zhou, Nana and Zhang, Lan and Yan, Hualin and Xu, Yi and Zhang, Zhongxin},
title = {Driver Fatigue Detection Based on Convolutional Neural Networks Using EM-CNN},
journal = {Computational Intelligence and Neuroscience},
volume = {2020},
number = {1},
pages = {7251280},


eprint = {https://onlinelibrary.wiley.com/doi/pdf/10.1155/2020/7251280},
year = {2020}
}

@article{minhas2022smart,
  author = {Minhas, Abid Ali and Jabbar, Sohail and Farhan, Muhammad and ul Islam, Muhammad Najam},
  title = {A smart analysis of driver fatigue and drowsiness detection using convolutional neural networks},
  journal = {Multimedia Tools and Applications},
  volume = {81},
  number = {19},
  pages = {26969--26986},
  year = {2022}
}

@article{han2025dense,
  author = {Han, Qing and Cui, Shimiao and Min, Weidong and Yan, Cong and Liu, Li and Ning, Feng and Li, Li},
  title = {A dense multi-pooling convolutional network for driving fatigue detection},
  journal = {Scientific Reports},
  volume = {15},
  number = {1},
  pages = {15518},
  year = {2025}
}

@ARTICLE{9557757,
  author={Bai, Jing and Yu, Wentao and Xiao, Zhu and Havyarimana, Vincent and Regan, Amelia C. and Jiang, Hongbo and Jiao, Licheng},
  journal={IEEE Transactions on Cybernetics}, 
  title={Two-Stream Spatial–Temporal Graph Convolutional Networks for Driver Drowsiness Detection}, 
  year={2022},
  volume={52},
  number={12},
  pages={13821-13833}
}

@inproceedings{tran2015learning,
  author = {Tran, Du and Bourdev, Lubomir and Fergus, Rob and Torresani, Lorenzo and Paluri, Manohar},
  title = {Learning Spatiotemporal Features with 3D Convolutional Networks},
  booktitle = {Proceedings of the IEEE International Conference on Computer Vision},
  pages = {4489--4497},
  year = {2015}
}

@inproceedings{carreira2017quo,
  author = {Carreira, Joao and Zisserman, Andrew},
  title = {Quo Vadis, Action Recognition? A New Model and the Kinetics Dataset},
  booktitle = {Proceedings of the IEEE Conference on Computer Vision and Pattern Recognition},
  pages = {4724--4733},
  year = {2017}
}

@ARTICLE{9185000,
  author={Huang, Rui and Wang, Yan and Li, Zijian and Lei, Zeyu and Xu, Yufan},
  journal={IEEE Transactions on Intelligent Transportation Systems}, 
  title={RF-DCM: Multi-Granularity Deep Convolutional Model Based on Feature Recalibration and Fusion for Driver Fatigue Detection}, 
  year={2022},
  volume={23},
  number={1},
  pages={630-640}
}

@article{wijnands2020realtime,
  author = {Wijnands, Jasper S. and Thompson, Jason and Aschwanden, Gideon D. A. and Stevenson, Mark},
  title = {Real-time monitoring of driver drowsiness on mobile platforms using 3D neural networks},
  journal = {Neural Computing and Applications},
  volume = {32},
  number = {13},
  pages = {9731--9743},
  year = {2020},
  publisher = {Springer}
}

@article{JIANG2023119640,
title = {Face2Nodes: Learning facial expression representations with relation-aware dynamic graph convolution networks},
journal = {Information Sciences},
volume = {649},
pages = {119640},
year = {2023},
issn = {0020-0255},


author = {Fan Jiang and Qionghao Huang and Xiaoyong Mei and Quanlong Guan and Yaxin Tu and Weiqi Luo and Changqin Huang}}

@article{al2023lstm,
  title={LSTM inefficiency in long-term dependencies regression problems},
  author={Al-Selwi, Safwan Mahmood and Hassan, Mohd Fadzil and Abdulkadir, Said Jadid and Muneer, Amgad and others},
  journal={Journal of Advanced Research in Applied Sciences and Engineering Technology},
  volume={30},
  number={3},
  pages={16--31},
  year={2023}
}

@article{cao2025optimized,
  title={Optimized driver fatigue detection method using multimodal neural networks},
  author={Cao, Shengli and Feng, Peihua and Kang, Wei and Chen, Zeyi and Wang, Bo},
  journal={Scientific Reports},
  volume={15},
  number={1},
  pages={12240},
  year={2025},
  publisher={Nature Publishing Group UK London}
}

@article{Wang2023Selective,
title={Selective Structured State-Spaces for Long-Form Video Understanding},
author={Jue Wang and Wenjie Zhu and Pichao Wang and Xiang Yu and Linda Liu and Mohamed Omar and Raffay Hamid},
journal={2023 IEEE/CVF Conference on Computer Vision and Pattern Recognition (CVPR)},
year={2023},
pages={6387-6397}
}

@inproceedings{10.1007/978-3-031-19833-5_6,
author = {Islam, Md Mohaiminul and Bertasius, Gedas},
title = {Long Movie Clip Classification with State-Space Video Models},
year = {2022},
isbn = {978-3-031-19832-8},
publisher = {Springer-Verlag},
address = {Berlin, Heidelberg},


pages = {87–104},
numpages = {18},
location = {Tel Aviv, Israel}
}

@article{2025Rethinking,
title={Rethinking the long-range dependency in Mamba/SSM and transformer models},
author={Cong Ma and K. Najarian},
journal={ArXiv},
year={2025},
volume={abs/2509.04226}
}

@article{Tang2024VMRNN,
title={VMRNN: Integrating Vision Mamba and LSTM for Efficient and Accurate Spatiotemporal Forecasting},
author={Yujin Tang and Peijie Dong and Zhenheng Tang and Xiaowen Chu and Junwei Liang},
journal={2024 IEEE/CVF Conference on Computer Vision and Pattern Recognition Workshops (CVPRW)},
year={2024},
pages={5663-5673}
}

@misc{somvanshi2025s4mambacomprehensivesurvey,
      title={From S4 to Mamba: A Comprehensive Survey on Structured State Space Models}, 
      author={Shriyank Somvanshi and Md Monzurul Islam and Mahmuda Sultana Mimi and Sazzad Bin Bashar Polock and Gaurab Chhetri and Subasish Das},
      year={2025},
      eprint={2503.18970},
      archivePrefix={arXiv},
      primaryClass={cs.LG}
}

@misc{lugaresi2019mediapipeframeworkbuildingperception,
      title={MediaPipe: A Framework for Building Perception Pipelines}, 
      author={Camillo Lugaresi and Jiuqiang Tang and Hadon Nash and Chris McClanahan and Esha Uboweja and Michael Hays and Fan Zhang and Chuo-Ling Chang and Ming Guang Yong and Juhyun Lee and Wan-Teh Chang and Wei Hua and Manfred Georg and Matthias Grundmann},
      year={2019},
      eprint={1906.08172},
      archivePrefix={arXiv},
      primaryClass={cs.DC}
}

@article{Feng_You_Zhang_Ji_Gao_2019, 
title={Hypergraph Neural Networks}, 
volume={33}, 
 
 
number={01}, 
journal={Proceedings of the AAAI Conference on Artificial Intelligence}, 
author={Feng, Yifan and You, Haoxuan and Zhang, Zizhao and Ji, Rongrong and Gao, Yue}, 
year={2019}, 
month={Jul.}, 
pages={3558-3565} 
}

@article{10.5555/1577069.1755843,
author = {King, Davis E.},
title = {Dlib-ml: A Machine Learning Toolkit},
year = {2009},
issue_date = {12/1/2009},
publisher = {JMLR.org},
volume = {10},
issn = {1532-4435},
journal = {J. Mach. Learn. Res.},
month = dec,
pages = {1755–1758},
numpages = {4}
}

@misc{gu2024mambalineartimesequencemodeling,
      title={Mamba: Linear-Time Sequence Modeling with Selective State Spaces}, 
      author={Albert Gu and Tri Dao},
      year={2024},
      eprint={2312.00752},
      archivePrefix={arXiv},
      primaryClass={cs.LG}
}

@article{Gu2021EfficientlyML,
  title={Efficiently Modeling Long Sequences with Structured State Spaces},
  author={Albert Gu and Karan Goel and Christopher R'e},
  journal={ArXiv},
  year={2021},
  volume={abs/2111.00396}
}

@data{e1qm-hb90-20,


author = {Shabnam Abtahi and Mona Omidyeganeh and Shervin Shirmohammadi and Behnoosh Hariri},
publisher = {IEEE Dataport},
title = {YawDD: Yawning Detection Dataset},
year = {2020} }

@misc{ghoddoosian2019realisticdatasetbaselinetemporal,
      title={A Realistic Dataset and Baseline Temporal Model for Early Drowsiness Detection}, 
      author={Reza Ghoddoosian and Marnim Galib and Vassilis Athitsos},
      year={2019},
      eprint={1904.07312},
      archivePrefix={arXiv},
      primaryClass={cs.CV}
}

@article{Yang2022FatigueView,
author = {Yang, Cong and Yang, Zhenyu and Li, Weiyu and See, John},
journal = {IEEE Transactions on Intelligent Transportation Systems},
title = {FatigueView: A Multi-Camera Video Dataset for Vision-based Drowsiness Detection},
year = {2023},
volume = {24},
number = {1},
pages = {233-246}
}

@inproceedings{NEURIPS2022_416f9cb3,
 author = {Tong, Zhan and Song, Yibing and Wang, Jue and Wang, Limin},
 booktitle = {Advances in Neural Information Processing Systems},
 editor = {S. Koyejo and S. Mohamed and A. Agarwal and D. Belgrave and K. Cho and A. Oh},
 pages = {10078--10093},
 publisher = {Curran Associates, Inc.},
 title = {VideoMAE: Masked Autoencoders are Data-Efficient Learners for Self-Supervised Video Pre-Training},
 
 volume = {35},
 year = {2022}
}

@ARTICLE{10382460,
  author={Yang, Lie and Yang, Haohan and Wei, Henglai and Hu, Zhongxu and Lv, Chen},
  journal={IEEE Transactions on Intelligent Transportation Systems}, 
  title={Video-Based Driver Drowsiness Detection With Optimised Utilization of Key Facial Features}, 
  year={2024},
  volume={25},
  number={7},
  pages={6938-6950}
}

@ARTICLE{9618813,
  author={Mou, Luntian and Zhou, Chao and Xie, Pengtao and Zhao, Pengfei and Jain, Ramesh and Gao, Wen and Yin, Baocai},
  journal={IEEE Transactions on Multimedia}, 
  title={Isotropic Self-Supervised Learning for Driver Drowsiness Detection With Attention-Based Multimodal Fusion}, 
  year={2023},
  volume={25},
  number={},
  pages={529-542}
}

@INPROCEEDINGS{11247563,
  author={Ren, Jing and Ma, Suyu and Jia, Hong and Xu, Xiwei and Lee, Ivan and Fayek, Haytham and Li, Xiaodong and Xia, Feng},
  booktitle={2025 IEEE/RSJ International Conference on Intelligent Robots and Systems (IROS)}, 
  title={LiteFat: Lightweight Spatio-Temporal Graph Learning for Real-Time Driver Fatigue Detection}, 
  year={2025},
  volume={},
  number={},
  pages={8059-8066}
}

@InProceedings{10.1007/978-3-030-66823-5_23,
author="Ortega, Juan Diego
and Kose, Neslihan
and Ca{\~{n}}as, Paola
and Chao, Min-An
and Unnervik, Alexander
and Nieto, Marcos
and Otaegui, Oihana
and Salgado, Luis",
editor="Bartoli, Adrien
and Fusiello, Andrea",
title="DMD: A Large-Scale Multi-modal Driver Monitoring Dataset for Attention and Alertness Analysis",
booktitle="Computer Vision -- ECCV 2020 Workshops",
year="2020",
publisher="Springer International Publishing",
address="Cham",
pages="387--405",
isbn="978-3-030-66823-5"
}

@INPROCEEDINGS{9008780,
  author={Feichtenhofer, Christoph and Fan, Haoqi and Malik, Jitendra and He, Kaiming},
  booktitle={2019 IEEE/CVF International Conference on Computer Vision (ICCV)}, 
  title={SlowFast Networks for Video Recognition}, 
  year={2019},
  volume={},
  number={},
  pages={6201-6210}
}

@misc{sun2023retentivenetworksuccessortransformer,
      title={Retentive Network: A Successor to Transformer for Large Language Models}, 
      author={Yutao Sun and Li Dong and Shaohan Huang and Shuming Ma and Yuqing Xia and Jilong Xue and Jianyong Wang and Furu Wei},
      year={2023},
      eprint={2307.08621},
      archivePrefix={arXiv},
      primaryClass={cs.CL}
}

@inproceedings{NEURIPS2021_bcc0d400,
 author = {Wu, Haixu and Xu, Jiehui and Wang, Jianmin and Long, Mingsheng},
 booktitle = {Advances in Neural Information Processing Systems},
 editor = {M. Ranzato and A. Beygelzimer and Y. Dauphin and P.S. Liang and J. Wortman Vaughan},
 pages = {22419--22430},
 publisher = {Curran Associates, Inc.},
 title = {Autoformer: Decomposition Transformers with Auto-Correlation for Long-Term Series Forecasting},
 
 volume = {34},
 year = {2021}
}

@inproceedings{zhou2021informer,
  title={Informer: Beyond Efficient Transformer for Long Sequence Time-Series Forecasting},
  author={Zhou, Haoyi and Zhang, Shanghang and Peng, Jieqi and Zhang, Shuai and Li, Jianxin and Xiong, Hui and Zhang, Wancai},
  booktitle={Proceedings of the AAAI Conference on Artificial Intelligence},
  volume={35},
  number={12},
  pages={11106--11115},
  year={2021}
}

@inproceedings{NEURIPS2024_baa2da9a,
 author = {Liu, Yue and Tian, Yunjie and Zhao, Yuzhong and Yu, Hongtian and Xie, Lingxi and Wang, Yaowei and Ye, Qixiang and Jiao, Jianbin and Liu, Yunfan},
 booktitle = {Advances in Neural Information Processing Systems},
 
 editor = {A. Globerson and L. Mackey and D. Belgrave and A. Fan and U. Paquet and J. Tomczak and C. Zhang},
 pages = {103031--103063},
 publisher = {Curran Associates, Inc.},
 title = {VMamba: Visual State Space Model},
 
 volume = {37},
 year = {2024}
}

@article{An2025ASH,
  title={A spatiotemporal hypergraph self-attention neural networks framework for the identification and pharmacological efficacy assessment of Parkinson’s disease motor symptoms},
  author={Xiaopeng An and Lu Su and Qi Yang and Bo Shen and Linhua Gan and Jia jun Ji and Jian Wang and Haifeng Su},
  journal={NPJ Parkinson's Disease},
  year={2025},
  volume={11}
}

@article{Wang2026MSTHGCNAM,
  title={MST-HGCN: A multimodal spatio-temporal hypergraph convolutional network for infantile spasms detection},
  author={Yi Wang and Haoran Luo and Luyang Meng and Yuying Fan},
  journal={Journal of King Saud University Computer and Information Sciences},
  year={2026}
}

@Article{e26030239,
AUTHOR = {Fan, Zunguan and Feng, Yifan and Wang, Kang and Li, Xiaoli},
TITLE = {Multi-Modal Temporal Hypergraph Neural Network for Flotation Condition Recognition},
JOURNAL = {Entropy},
VOLUME = {26},
YEAR = {2024},
NUMBER = {3},
ARTICLE-NUMBER = {239},
PubMedID = {38539751},
ISSN = {1099-4300}
}

@article{Wang2024HyperSTTNSG,
  title={Hyper-STTN: Social Group-aware Spatial-Temporal Transformer Network for Human Trajectory Prediction with Hypergraph Reasoning},
  author={Weizheng Wang and Le Mao and Baijian Yang and Guohua Chen and Byung-Cheol Min},
  journal={ArXiv},
  year={2024},
  volume={abs/2401.06344}
}
% \appendix









\end{document}